\documentclass[journal]{IEEEtran}
\usepackage[switch]{lineno}
\ifCLASSINFOpdf
\else
\fi
\usepackage{url}
\usepackage[T1]{fontenc}
\usepackage{times}
\usepackage{latexsym}
\usepackage{multirow}
\usepackage{multicol}
\usepackage{booktabs}
\usepackage{enumitem}
\usepackage[utf8]{inputenc}
\usepackage[scaled=0.8]{beramono}
\usepackage{microtype}
\usepackage{amsmath}

\usepackage{xcolor}
\usepackage[normalem]{ulem}

\usepackage{graphicx}
\usepackage{tikz}
\usepackage{forest}
\usetikzlibrary{trees,positioning,shapes,shadows,arrows.meta}

\begin{document}
%
\title{Multilingual Text Representation}
%
%
%

\author{Fahim Faisal\\
        Department of Computer Science\\
        George Mason University\\
        ffaisal@gmu.edu}

\maketitle

\begin{abstract}
Modern NLP breakthrough includes large multilingual models capable of performing tasks across more than 100 languages. State-of-the-art language models came a long way, starting from the simple one-hot representation of words capable of performing tasks like natural language understanding, common-sense reasoning, or question-answering, thus capturing both the syntax and semantics of texts. At the same time, language models are expanding beyond our known language boundary, even competitively performing over very low-resource dialects of endangered languages. However, there are still problems to solve to ensure an equitable representation of texts through a unified modeling space across language and speakers. In this survey, we shed light on this iterative progression of multilingual text representation and discuss the driving factors that ultimately led to the current state-of-the-art. Subsequently, we discuss how the full potential of language democratization could be obtained, reaching beyond the known limits and what is the scope of improvement in that space. 
\end{abstract}


%
\IEEEpeerreviewmaketitle

\section{Introduction}
Natural language processing (NLP) primarily involves making linguistic-specific applications for machines to understand language. Earlier days of NLP development mainly focused on the idea of distributional hypothesis~\cite{harris54} that is, \textit{"words occurring in the same context tend to have a similar meaning or closely related meaning"}. From there, NLP has come a long way in the modeling language. The tasks NLP tries to solve are complex and multidimensional if we put them into the perspective of the machine and numerical mapping. For example, there are multiple dynamics to deal with here, like various languages and dialects to consider and tasks to solve with different levels of granularity. Combining all these dynamics in a unified representation space is a complex problem. The base starting point could be just words, as the word is one such unit of language that is quite universal. Following this thought, the most straightforward idea could be to compute the word frequency, thus constructing a count-based numerical mapping that we can think of as a computationally feasible language representation. N-gram language models are representations that have led the domain of NLP for a substantial time. Later on, advancing over the distributional scheme of language representation coupled with the availability of huge computation resources, neural models became the go-to approach for all kinds of NLP tasks. Researchers investigated different theories and directions in this domain of neural language modeling before transformer-based neural networks revolutionized NLP. The transformer-based model provides exceptional text representation utilizing the multi-layered encoder blocks~\cite{https://doi.org/10.48550/arxiv.1706.03762}. This is useful as text gets different meanings based on how it is used in a context. In addition, multiple languages can share a single representation space using transformers. This effectively led to the path for multilingual text representation, where data collected from languages existing all over the world can be accumulated in a single setting, and models can learn and perform actionable inference on a wide variety of tasks comprising language and dialectal varieties. Though a monolingual or region-focused transformer still vastly outperforms a more generalized multilingual model on most tasks, it is not always feasible to train multiple versions of the domain-focused model. The idea is to make a shared representation space that effectively works for many languages, while the resource scarcity of specific languages should benefit from other high-resource languages. mBERT~\cite{devlin-etal-2019-bert} and xlm-r~\cite{conneau-etal-2020-unsupervised} trained with multiple objective functions on more than hundreds of languages came a long way in achieving this vision. However, the full potential of a unified multilingual text representation is still a significant research problem to solve. Because, quite regularly, the inclusion of new tasks and languages in the modeling paradigm points out the fact that, when these models move beyond the monolingual scheme, the total capacity of the model gets distributed across languages, thus often resulting in capacity dilution~\cite{wang-etal-2020-negative}. An ideal scenario would be to have no amount of negative interference, such that we get an equitable performance across languages. Another important direction for multilingual models is to ensure the easy-expand-ability to new languages and adaptability to new tasks. 

Keeping all these advanced development of multilingual text representation in context, in this survey, we provide insight into the open problems and questions to look for. In addition, we discuss how the text representation model starting from the count-based vector representation of words, came to the point of a multilingual text representation model capable of performing across more than 100 languages. We structure the contents based on the following contexts: (1) How did the text representation model make the iterative progress, and which were the driving factors in each step? (2) What are the primary building blocks of a unified multilingual text representation model, and how do they vary given the difference in scenarios? (3) What are the current barriers that limit the progression of full-scale multilingual text representation? (4) The fairness and interpretability of currently available models and how equitable they serve the intended user’s utility.

\section{Preliminaries}

\subsection{Terminologies}

\paragraph{Tokenization}: A process of transforming the text into tokens of words is generally known as tokenization \cite{BALAZS201695}. Not necessarily; it needs to be in words, but it can be divided into words, symbols, phrases, or sub-word tokens.

\paragraph{Word segmentation} Separating the phrase, content, and keywords. Other steps include stemming, lemmatizing, handling negation, and separating punctuation.

\paragraph{Embedding} Embedding is a relatively low-dimensional space where we can place the transformed high-dimensional vectors. Word embedding generally means a type of embedding space where the close-meaning words would be grouped, maintaining a close distance. 

\paragraph{Attention}

We can interpret this mechanism as the vector of important weights~\cite{https://doi.org/10.48550/arxiv.1409.0473}. An attention vector provides insight into how words in a sentence might be correlated with other words in that sentence. In addition, it determines the most useful blocks of the input in terms of transferring the contextual insights to actionable information for the target output. The attention mechanism mitigates the long-lasting disadvantage of fixed-length vectors in sequence-to-sequence models: the incapability of remembering long sentences. 

\subsection{Early days of Multilingual Word Representation}
Multilingual Word representation involves representing words from different languages through a shared space. The bilingual word embedding model~\cite{zou-etal-2013-bilingual} is one of the earlier works that uses machine translation word alignment and embeds words from the source and target language into a single vector space. During training, it imposes constraints over the distance based on the translation pairs of two languages. Later on, \cite{https://doi.org/10.48550/arxiv.1309.4168} proposed an approach that can learn bilingual word embedding given the lexicon of bilingual word pairs. It performs a linear transformation to transform the source language vectors to the target language vector space. However, when provided a lexicon of small size, this model performs poorly. To tackle this issue, \cite{10.5555/3298023.3298059} introduced a matching score mechanism along with the original bilingual lexicon pairs to bring the embeddings closer.

\subsection{Neural Language Model Basics}
In neural language modeling (NLM), we train a probabilistic classifier to predict a probability distribution where the conditional probability of selecting word $w_i$ is learned using various kinds of neural networks (e.g., feed-forward, recurrent, etc.):
\begin{align*}
    P(s) = \prod_{i=1}^l P(w_i|w_1^{i-1})
\end{align*}

\paragraph{Feed-forward NLM:} A feed-forward NLM (FFN)~\cite{10.5555/944919.944966} adopts the notion of an n-gram language model by assuming each word in a sequence depends on the words closer to it statistically though it fails to consider the long-term dependency. Instead of considering the dependence of the whole previous sequence, a context window ($i-n+1\, to\, i-1$) is used for better approximation: $P(w_i|W_1^{i-1} = P(w_i|W_{i-n+1}^{i-1})$. The context word sequence $x=[w_{i-n+1}...,w_{i-1}]$ is fed into a FFN, and later, a softmax layer over the final output matrix is used to get the output probability of $P(w_i|W_{i-n}^{i-1})$.

\paragraph{Convolutional NLM:} This one enhances the ffn by injecting a CNN layer on top of the input representation~\cite{pham-etal-2016-convolutional}, which involves a sliding window of the input vectors centered on each word vector and later on, performing max-pooling on it.

\paragraph{Recurrent NLM} Recurrent Neural Network (RNN) based LM~\cite{MikolovKBCK10} addresses the issue of long-term dependency problem. At every time step of RNN, the input is just the previous word vector instead of the concatenated vectors of n previous words. Meanwhile, the information of all previous words is preserved by the internal state of RNN. The most common RNN types are LSTM~\cite{10.1162/neco.1997.9.8.1735} and GRU~\cite{cho-etal-2014-properties}. The key problem with RNN-based language models is falling into the vanishing gradient (taking the multiplication of a large number of derivatives eventually results in a value close to zero, which can further not be used in error function calculation). This failure leads to the problem of not capturing the dependencies among words in a long sentence as the amount of computation using RNN increases when the distance among words increase in a sentence.

\paragraph{Transformer Language Models}
This is a non-recurrent encoder-decoder architecture with a series of attention-based blocks. An encoder prepares a contextual representation of the given input, whereas the decoder can generate output based on the output segments that are already generated. The attention mechanism is at the core of the Transformer architecture~\cite{https://doi.org/10.48550/arxiv.1706.03762}. An encoder-decoder-based transformer contains three types of attention units facilitated with the help of queries, keys, and values.

Key $K$: This is a label of a word and is used to distinguish between different words.

Query $Q$: It represents an active request that checks all the keys and selects the one that matches the request.

Value $V$: A value is always paired with a key, and when the query selects a key, its value is the one that propagates. A value is an information a word contains.

When $k, Q,\&V$ all get generated from the same source sequence, it is self-attention (exist one in the encoder, one in the decoder). When they come from different sequences, they form a cross-attention mechanism (happens in between encoder-decoder interactions). The other type is the scaled-dot product attention mechanism that performs dot product calculation between $Q$ and $K$ matrices. Before that, the $k, Q,\&V$ are calculated by the matrix multiplication with learned word vectors. The scaled-dot product attention then ensures that we select more information from the values where the key and the query are more similar. On top of these mechanisms, we have multi-head attention where multiple scaled dot product attentions run in parallel, thus helping the network to attend to multiple pieces of information simultaneously. In addition, there is no recurrent element in transformers. So, to learn the position information of each word, a combination of sine and cosine waves of different frequencies is used, as each position would have a unique combination of values. There are other essential segments like residual connections and layer normalization. Residual or skip connection adds the input to the output after a layer, allowing the gradients to flow directly through the network. Whereas the layer normalization keeps the mean of each training sample close to 0 and the standard deviation close to 1, thus stabilizing the training as well as reducing the training time.

In practice, a layer from the encoder contains multi-head attention sub-layers and a position-wise feed-forward sub-layer. The sub-layers are connected using the residual connections. The multi-head attention sub-layer contains several attention heads. A head is a scaled dot product attention structure taking the query matrix $Q$, the key matrix $k$, and the value matrix $V$. So the output,
\begin{align*}
    Attention(Q,K,V) &= Softmax(\frac{QK_T}{\sqrt{d_k}})V\\
    Multilhead(Q,K,V) &= [head_1,head_2,..head_h]W^0\\
    \text{where, }head_i &= Attention(QW_i^Q, KW_i^K,VW_i^V)\\
    \text{and }W_i^Q &= \text{linear projections},\\
    \text{and }d_k &= \text{is the dimension of the query matrix.}
\end{align*}

The fully connected position-wise feed-forward sub-layers contain two linear transformations with RELU:
\begin{align*}
    FFN(x) &= W_2 max(0, W_1x+b_1)+b_2
\end{align*}
Transformers are better for dealing with long-term dependency than RNNs. Originally it was proposed to solve the problem of machine translation, but later on, it became the backbone of all kinds of NLP applications.

\subsection{Multilingual Language Modeling}
A general Multilingual Language Model (MLLM) architecture contains an input layer, transformer layer, and output layer.
\begin{itemize}
    \item input layer: sequence of tokens from a representation of one-hot subword token vocabulary (concatenated from all languages).
    \item transformer layer: each layer contains k attention heads, followed by a feed-forward neural network.
    \item output layer: contextual representation for each token. It contains a simple linear transformation followed by a Softmax that takes the last layer token representation and produces the probability distribution.
\end{itemize}

\section{Area Taxonomy}

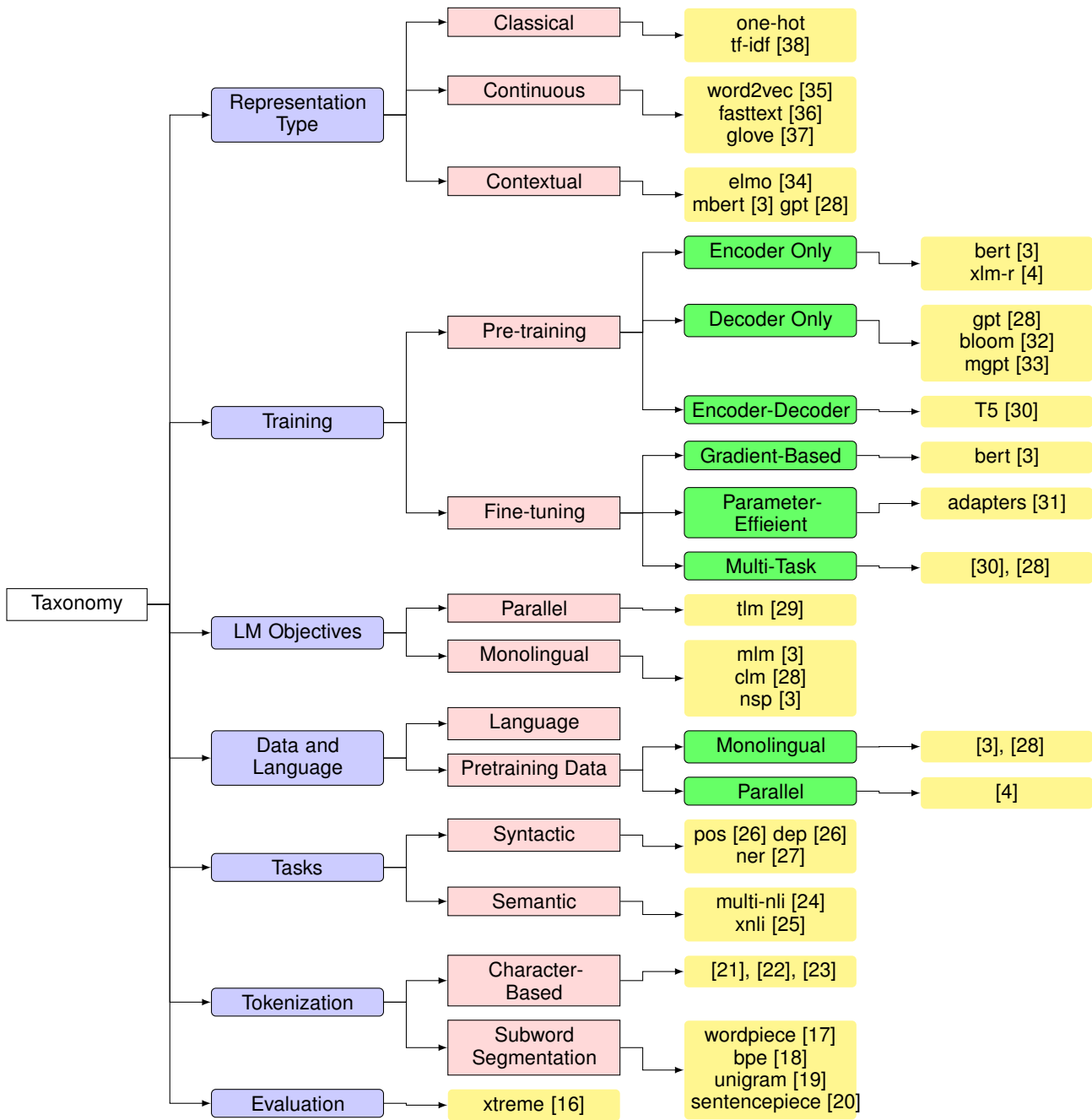
\begin{figure*}
\small
    \centering

\tikzset{
    basic/.style  = {draw, text width=2cm, align=center, font=\sffamily, rectangle},
    root/.style   = {basic, rounded corners=2pt, thin, align=center, fill=green!30},
    onode/.style = {basic, thin, rounded corners=2pt, align=center, fill=green!60,text width=2.5cm,},
    tnode/.style = {basic, thin, align=left, fill=pink!60, text width=2.5cm, align=center},
    xnode/.style = {basic, thin, rounded corners=2pt, align=center, fill=blue!20,text width=2.5cm,},
    wnode/.style = {basic, thin, align=left, fill=pink!10!blue!80!red!10, text width=6.5em},
    cnode/.style = {basic, thin, align=center, rounded corners=1pt, fill=yellow!110!blue!50, rounded corners=2pt,draw=none, text width=2.5cm},
    edge from parent/.style={draw=black, edge from parent fork right}

}
\hspace{-2em}
\begin{forest} for tree={
    grow=east,
    growth parent anchor=west,
    parent anchor=east,
    child anchor=west,
    edge path={\noexpand\path[\forestoption{edge},->, >={latex}] 
         (!u.parent anchor) -- +(10pt,0pt) |-  (.child anchor) 
         \forestoption{edge label};}
}
[Taxonomy, basic,  l sep=10mm,
    [Evaluation, xnode,  l sep=10mm,
        [xtreme~\cite{hu2020xtreme}, cnode]
    ]
    [Tokenization, xnode,  l sep=10mm,
        [Subword\\Segmentation, tnode , l sep=10mm,
            [wordpiece~\cite{6289079}\\bpe~\cite{sennrich-etal-2016-neural}\\unigram~\cite{https://doi.org/10.48550/arxiv.1804.10959}\\sentencepiece~\cite{kudo-richardson-2018-sentencepiece}, cnode]
        ]
        [Character-\\Based, tnode,  l sep=10mm,
            [~\cite{huang2013learning,Hwang2017CharacterlevelLM,clark-etal-2022-canine}, cnode]
        ]
    ]
    [Tasks, xnode,  l sep=10mm,
        [Semantic, tnode , l sep=10mm,
            [multi-nli~\cite{multi-nli} xnli~\cite{conneau-etal-2018-xnli}, cnode]
        ]
        [Syntactic, tnode,  l sep=10mm,
            [pos~\cite{udp} dep~\cite{udp} ner~\cite{rahimi-etal-2019-massively}, cnode]
        ]
    ]
    [Data and Language, xnode,  l sep=10mm,
        [Pretraining Data, tnode,  l sep=10mm,
            [Parallel, onode, l sep=10mm,
                [\cite{conneau-etal-2020-unsupervised}, cnode]
            ]
            [Monolingual, onode , l sep=10mm,
                [~\cite{devlin-etal-2019-bert, brown2020language}, cnode]
            ]
        ]
        [Language, tnode]
    ]
    [LM Objectives, xnode,  l sep=10mm,
        [Monolingual, tnode , l sep=10mm,
            [mlm~\cite{devlin-etal-2019-bert}\\clm~\cite{brown2020language}\\nsp~\cite{devlin-etal-2019-bert}, cnode]
        ]
        [Parallel, tnode, l sep=10mm,
            [tlm~\cite{conll-2018-conll}, cnode]
        ]
    ]
    [Training, xnode,  l sep=10mm,
        [Fine-tuning, tnode,  l sep=10mm,
            [Multi-Task, onode, l sep=10mm,
                [~\cite{raffel2019exploring,brown2020language}, cnode]
            ]
            [Parameter-Effieient, onode, l sep=10mm,
                [adapters \cite{pfeiffer-etal-2020-mad}, cnode]
            ]
            [Gradient-Based, onode, l sep=10mm,
                [bert~\cite{devlin-etal-2019-bert}, cnode]
            ]
        ]
        [Pre-training, tnode,  l sep=10mm,
            [Encoder-Decoder, onode, l sep=10mm,
                [T5~\cite{raffel2019exploring}, cnode]
            ]
            [Decoder Only, onode, l sep=10mm,
                [gpt~\cite{brown2020language} bloom~\cite{bloom} mgpt~\cite{https://doi.org/10.48550/arxiv.2204.07580}, cnode]
            ]
            [Encoder Only, onode, l sep=10mm,
                [bert~\cite{devlin-etal-2019-bert}\\xlm-r~\cite{conneau-etal-2020-unsupervised}, cnode]
            ]
        ]
    ]
    [Representation Type, xnode,  l sep=10mm,
        [Contextual, tnode, l sep=10mm,
            [elmo~\cite{peters-etal-2018-deep} mbert~\cite{devlin-etal-2019-bert} gpt~\cite{brown2020language}, cnode]
        ]
        [Continuous, tnode, l sep=10mm,
            [word2vec~\cite{word2vec} fasttext~\cite{bojanowski-etal-2017-enriching} glove~\cite{pennington-etal-2014-glove}, cnode]
        ]
        [Classical, tnode, l sep=10mm,
            [one-hot\\tf-idf~\cite{ref1}, cnode]
        ]
    ]
]
\end{forest}

    \caption{Area Taxonomy of Multilingual Text Representation}
    \label{fig:lit_surv}
\end{figure*}

The multilingual text representation area taxonomy is presented in Figure \ref{fig:lit_surv}. Here we divide the taxonomy into certain primary parts, thus representing the building blocks of text representation from different perspectives. For example, \textit{Representation Type} defines ways of representing text starting from the classical models to recent Transformer breakthroughs. Furthermore, we can divide a standard NLP training paradigm into parts like pre-training and fine-tuning. We believe tokenization, being a step from text preprocessing, deserves a separate branch of discussion because of its impact on multilingual settings. In the next section, we report a detailed discussion of the existing research in every direction of this taxonomy.

\section{Taxonomy-Based Survey}
\subsection{Representation Type} The simplest form of text representation would be the word-count-based one that fully relies on statistical information. From there, text representation has advanced to a label, where we can represent texts from different languages in a shared space. The progression is described in detail here.

\paragraph{Classical Models}
Word frequency is the basis of the classical text representation model used in earlier days. We can divide these models into two parts: (1) Categorical and (2) Weighted. One-hot-encoding and Bag-of-Words (BoW) are the categorical models. In one-hot encoding, the dimension to represent texts is equal to the terms present in the vocabulary, where binary values are used to define the presence or absence of a particular term. Whereas, BoW is just an updated one-hot-encoding where we sum all the one-hot-representation of words in a sentence. However, these categorical models fail to capture semantic relations and the order of the words. A weighted text representation model known as Tf-Idf was introduced to solve this problem. A Tf (i.e., term-frequency) matrix just divides the word count by the length of a document, thus identifying how often a word occurs in a document. Whereas, Idf (i.e., inverse document frequency) matrix tries to reduce the effect of common words by putting more weights on the critical words (words that are not equally frequent in all documents, like stop words).
 
\paragraph{Continuous Representation}
One popular approach to represent text is to present it as vectors where each dimension corresponds to the frequency of words, thus resulting in a word vector. The pitfall is that the vector length might be substantial depending on the vocabulary size. In that case, adopting a dimensionality reduction procedure becomes a default choice. Though these reduced vectors might be compact and efficient to compute, they contain less of the original information. Moreover, the individual dimensions no longer preserve the interpretable features that could be mapped back to the original textual building blocks. In one way, the context gets distributed throughout the vector length, thus making it a distributed representation of continuous values. In this distributed/continuous representation, each dimension of a word type vector becomes a parameter to be learned and optimized based on the observable patterns in the data. We can see these parameters as continuous values that can be learned using a continuous objective function using iterative algorithms like gradient descent. Word2vec~\cite{word2vec} is one such distributed vector representation of text. Word2vec considers similar meaning words like \textit{"small"} and \textit{"smaller"} comparatively closer in the vector space. There are two types of word2vec algorithms in practice: (1) \textit{Continuous bag of words (CBOW)} (2) \textit{Skip-gram}. In CBOW, context is considered the input. The neural network tries to correlate the weight matrix with each word, thus improving the representation of words through backpropagation of the error gradient. In skip-gram, the context is estimated based on the given the word. However, both CBOW and Skip-gram were very time-consuming in practice. To solve this issue, Hierarchical softmax and negative sampling approaches were introduced. Negative sampling restricts the output sum so that only a subset of the vectors get updated in each step, whereas, in hierarchical softmax, words are chosen based on their count-based conditional probabilities. Later on, GLOVE~\cite{pennington-etal-2014-glove} improves significantly on word2vec by using the global contextual information by constructing the global co-occurrence matrix and factorizing it later on. However, these distributed representation methods failed to consider the out-of-vocabulary (OOV) words. This is when Fasttext~\cite{bojanowski-etal-2017-enriching} was proposed. Fasttext breaks the words to n-gram instead of using the full-word representation at once, thus solving the OOV issue. These models can extract syntactic and semantic information while dealing with specific corner cases. However, there is this existing issue of keeping the full context-specific representation of a document because understanding the actual context is required for most downstream tasks in NLP.

\paragraph{Contextual Word Vectors}
The weakness of the continuous word representation model was to failing to capture the global context information on a low-dimensional scale. To solve this issue, contextual word representation models like Context2vec~\cite{melamud-etal-2016-context2vec}, Cove~\cite{10.5555/3295222.3295377} and ELMO~\cite{Peters:2018} were proposed later on. These advanced models solved many of the existing issues but still face catastrophic forgetting.  Context2vec model is based on Word2vec’s CBOW model but replaces its average word representation within a fixed window with a better and more powerful Bi-directional LSTM neural network. Whereas Cove uses machine translation instead of the approach used in Word2vec (skip-gram or CBOW) or GloVe (Matrix factorization). They pre-train a two-layer BiLSTM for attending sequence-to-sequence translation, starting from GloVe word vectors. Then they took the output of the sequence encoder and called it a CoVe, followed by combining it with GloVe vectors, and used it in a downstream task-specific mode using transfer learning. On top of these models, ELMO (embeddings from language model)~\cite{peters-etal-2018-deep} was the one to successfully advance the contextualization of word vectors. The key idea was: while a word token will have its vector, this vector needs to depend on the nearby word contexts. This is similar to the distributional hypothesis~\cite{harris54}. However, unlike word-type vectors, these word token vectors combine the word-level vectors with neural network parameters going beyond the lookup table of word-type vectors. ELMO contains two neural networks: One for the left context (start of the sentence to the token) and another for the right context (from the token to the end of the sentence). It uses a recurrent neural network that was the most advanced neural network at that time and was a novel thing to introduce in language modelling. Later on, ULMFiT~\cite{howard-ruder-2018-universal} successfully adopt the concept of transfer-learning in the contextual text representation. It segments the end-to-end process into three steps then: (i) General LM pre-training, (ii) Target task LM fine-tuning, and (iii) Target task classifier fine-tuning. Then finally, the transformer-based fine-tuning appeared in the picture which proved to be more efficient and faster than LSTM or CNN for language modelling.

\paragraph{Transformer Based Language Models}
Transformer-based language models are more efficient than the ones with LSTM or CNN for language modeling. The first model to efficiently use the transformer architecture was GPT~\cite{brown2020language} which is a decoder-based model. GPT was trained with Causal Language Modeling. The next one to efficiently represent the semantics and context was BERT~\cite{devlin-etal-2019-bert}: Bidirectional Encoder Representations from Transformers. BERT uses a parallel attention layer instead of using sequential recurrent layers. It is an encoder model that uses masked language modelling (MLM) and next-sentence prediction (nsp) objectives in a self-supervised manner for being trained on a large pool of textual data. Here, the next sentence prediction is used to collect long-term or pragmatic information.

\subsection{Training}
We can largely divide the current practice of transformer-based model training into pre-training and fine-tuning. Pre-training is the training part of preparing a base language model that can further be transferred and adapted to any downstream task through fine-tuning.

\subsubsection{Pre-training} Pre-training is a common step in making large language models in the current NLP paradigm. Pre-training means training a language model on extensive textual data in a self-supervised manner. Self-supervised because the pre-training objective looks for the data labels to predict, which are automatically contracted from the data itself (e.g., masked language modeling, next sentence prediction). Here we discuss some common types of pretrained transformer-based models and their specific approaches to performing the training.

\paragraph{Decoder only} GPT~\cite{brown2020language} is the most common example of the decoder-only model. It uses the causal language modeling objective. Other advanced decoder only models are OPT~\cite{zhang2022opt}, BLOOM~\cite{bloom}, mGPT~\cite{https://doi.org/10.48550/arxiv.2204.07580}. Recently, these decoder-only models are also going multilingual (e.g., mGPT, BLOOM) and increasing the span of their parameters.

\paragraph{Encoder Only} The most common example of encoder only models are bert~\cite{devlin-etal-2019-bert}. It uses masked language modeling (MLM) and next sentence prediction (NSP) objectives to train on large monolingual texts. Several interesting works investigate improving the self-supervised pre-training objective of encoder-only models. For example, in~\cite{https://doi.org/10.48550/arxiv.1907.11692}, the authors find out that using dynamic masking during training(randomly masking 15\% of token each step) time instead of static masking improves the performance. In addition, they show that NSP can be dropped completely from the training objective to get better performance.

\paragraph{Encoder-Decoder} T5~\cite{raffel2019exploring} is a encoder-decoder model. The objective of T5 is closely related to the MLM and word dropout techniques. The difference with the original MLM is the consecutive span of corrupted or dropped tokens is replaced with one single sentinel token. The output sequence contains these dropped-out spans delimited by the sentinel tokens that replace the original text. Using this technique of denoising sequence-to-sequence objective, the decoder can predict the span of tokens in the masked position instead of just a single token.

\subsubsection{Fine-tuning}
Fine-tuning means adopting a pre-trained model for any downstream task. Previously, fine-tuning was only used to indicate the gradient-based training of the complete model. Here, we will consider any updates that make a base model further usable for any downstream task as fine-tuning.

\paragraph{Gradient-based}
The gradient-based finetuning means producing a whole set of new parameters by training the entire model and updating all parts for a downstream task. In the early days of transformer and monolingual language modeling, this was the go-to approach to try any model on a new task or language~\cite{devlin-etal-2019-bert}.

\paragraph{Parameter efficient}
The gradient-based finetuning is a massive bottleneck if we consider the number of downstream tasks. Then there is parameter-efficient finetuning named as adapters which include updating only parts of model parameters~\cite{pfeiffer-etal-2020-mad}. Adapters proved to help perform effective cross-lingual transfer while reducing the amount of negative interference~\cite{pfeiffer2022xmod}.

\paragraph{Multitask and zero-shot learning}
Multitask learning includes training the model on a wide variety of tasks simultaneously instead of just one task~\cite{raffel2019exploring}. Following this scheme, there would be less necessity for further finetuning. However, till now, the utilization of this scheme is still confined to only computationally intensive models and not in a general-purpose setting.

\subsection{Language Modeling Objectives}
We can think of the learning objectives of language modeling in a data-driven manner. One is monolingual objectives which generally work on monolingual data trying to figure out the representation of a missing token. Whereas in bilingual objectives, data comes from a parallel corpus. The aim is to represent similar-meaning words as close as possible.

\paragraph{Monolingual Objectives} Masked language modeling(MLM) is monolingual but surprisingly helps learn multilingual models. Consider, a sentence $X=(x_1,x_2,..x_s)$. In masked language modeling, token $x_i$ is replaced with $\bar{x_i}$. So the input to the model becomes $X=(x_1,...\bar{x_i},..x_s)$. Now the prediction task ins $Y=(x_i)$ where $x_i$ can be predicted from the final representation of $\bar{x_i}$~\cite{devlin-etal-2019-bert}. Previously it was thought that MLM works well because of its ability to discover syntactic and semantic mechanisms in the pre-training stage. However, recent findings suggest MLM learns the higher order distributional statistics, thus making it a very useful prior for further fine-tuning~\cite{sinha-etal-2021-masked}. Another common one is Causal Language Modeling (CLM). CLM is the traditional learning objective in a language model that includes estimating the probability of a word given the previous sequence of words in a sentence, thus $P(x_t|x_1,x_2..,x_{t-1},\theta)$. In other words, a model can be trained with inputs $X=(x_1,,,x_{t-1})$ and outputs $Y=(x_t)$. Here $x_t$ is the output label which can be predicted from the final layer representation of token $x_{t-1}$.

\paragraph{Parallel-corpora based Objectives} This is mainly a bilingual objective, where the model learns to reduce the distance of similar meaning text given a parallel corpus. As the amount of parallel corpus is much less than monolingual one, a joint objective utilizing both monolingual and parallel corpus is used. Translation language modeling (TLM)~\cite{NEURIPS2019_c04c19c2} is the most common one, where $x_1^A,x_2^A,...x_n^A$ is a sentence in language $A$ and $x_1^B,x_2^B,...x_m^B$ is a sentence in language $B$. Now the MLM masks tokens from both sentences $A$ and $B$. It tries to predict the missing token by utilizing either the surrounding tokens in $A$ or the translation in $B$. Another useful objective is to use the contextual representation from the base language model given word alignment~\cite{dou-neubig-2021-word, 10.3115/1072133.1072187, dyer-etal-2013-simple} between translations and ask the model explicitly to reduce the distance of similar meaning word representation~\cite{https://doi.org/10.48550/arxiv.2002.03518}.

\subsection{Data and Language}
\paragraph{Pre-training Data}
Multilingual Language Models explore two data types primarily for pre-training: (1) Large Monolingual Data and (2) Parallel Corpus. Bert~\cite{devlin-etal-2019-bert} uses monolingual data from Wikipedia, whereas XLM-R~\cite{conneau-etal-2020-unsupervised} uses a much larger common-crawl corpus to train the model. Language family or region-specific models~\cite{kakwani2020indicnlpsuite} have their focused source of data to explore.

\paragraph{Languages}
BERT~\cite{devlin-etal-2019-bert} and XLM-R~\cite{conneau-etal-2020-unsupervised} are two multilingual models having trained in more than 100 languages. One problem is that the data availability is not evenly distributed across languages. Usually, models use exponential smoothing to make the data ratio fair across high-resource and low-resource languages. The main idea is if a language $i$ contains $m\%$ of total pre-training data, then the probability of that language is $p_i=k/100$ where $p_i$ is exponentiated by a factor $\alpha$. Then the resulting values are normalized to give a probability value to all the languages. $\alpha < 1$ means the high-resourced ones will be under-sampled, whereas the low-resourced ones will be over-sampled.

\subsection{Tasks}
The range of mainstream NLP tasks requires models to perform transfer learning at different difficulty levels, thus requiring word, phrase, or sentence-level understanding. Essentially, we can frame the tasks into two main groups:
\paragraph{Syntactic Task} These tasks focus on the sentence-level or word-level structure of the languages. The most common example of this type of task includes dependency parsing (DEP)~\cite{https://doi.org/10.48550/arxiv.1611.01734}, named entity recognition (NER)~\cite{rahimi-etal-2019-massively} and parts of speech tagging (POS)~\cite{udp}. Universal dependency project~\cite{udp} contains a dataset of DEP and POS tasks covering more than 100 languages as part of the evaluation. 

\paragraph{Semantic Task}
Some tasks require models to perform language-level understanding or inference that can not be answered by just following the sentence structure. Natural language inference~\cite{multi-nli, conneau-etal-2018-xnli} that tries to predict whether a premise sentence entails, contradicts, or is neutral toward a hypothesis sentence is one such task.

\subsection{Tokenization}
Tokenization is splitting a sequence of characters given a document unit into a piece of singular units (e.g., tokens) based on some level of heuristics. The simplest form of heuristics would be to cut into words based on white space, thus preserving the meaning at the unit level. However, this comes with issues like phrase-level segmentation and different characterization of the similar-meaning word because of minor spelling variations. Moreover, words form differently in different languages. For example, in french, the use of apostrophe sometimes works like the mention of a definite article \textit{(l'ensemble)}. In German, compound words are written without white-space \textit{(computational modeling $\rightarrow$ Computermodellierung)}. Thus, researchers focus on sub-word and character-based splitting instead of white-space tokenization to perform a more usable segmentation. Tokenization does the text-to-numerical mapping of input, making it one of the primary steps before feeding a text distribution to a computation model. As languages can be of different scripts with different vocabulary, multilingual models accumulate all happening subwords as part of the vocabulary for the supporting languages. This becomes a limiting factor when we want to extend the model capacity to a thousand more languages. Training monolingual tokenizers each time is not a viable option, and the original multilingual tokenizer does not have vocabulary distributional knowledge about the unseen languages.

\paragraph{Subword Segmentation} Subword-based tokenizations are the most widely used tokenization for multilingual transformer-based models. Because simple whitespace-based tokenization suffers from the dimensionality bottleneck problem, simple character-based splitting results in losing all context signals. The main difference in common sub-word-based tokenizers lies in the choice of character pairs to merge. For example, BPE~\cite{sennrich-etal-2016-neural} makes a frequency-based merging, whereas the Unigram~\cite{https://doi.org/10.48550/arxiv.1804.10959} model uses a probability based merging (computing the likelihood of each subword instead of using the most frequent ones). Word-Piece~\cite{6289079} tries to utilize the advantage of both unigram and BPE. It merges based on likelihood instead of frequency, but the choice of words to join is based on frequency. SentencePiece~\cite{kudo-richardson-2018-sentencepiece} is another subword-based tokenizer that rules out the initial whitespace-based splitting (useful for languages like Chinese and Japanese) by considering space as just another character and then employing BPE or Ingram on top of that. In the multilingual scenario, languages can be of different scripts with different vocabulary, and multilingual models accumulate all happening subwords as part of the vocabulary from the supporting languages. Now subword-based tokenizer with a fixed vocabulary size performs unfairly for low-resource languages due to the data imbalance among languages resulting in excessive fragmentation of subwords~\cite{https://doi.org/10.48550/arxiv.2106.12672}. In addition, the fixed vocabulary becomes a limiting factor when we want to extend the model capacity to a thousand more languages.

\paragraph{Character-based Model}
Subword tokenizers eliminate the out-of-vocabulary problem to a large extent, though the reliance on static vocabulary prevents end-to-end learning across all languages. One alternative would be to use a character-based approach~\cite{huang2013learning,Hwang2017CharacterlevelLM}. Though this is more adaptable to the code-switched language and noisy text, they may not capture the token-level representation and convert the longer sequence to character-level representation~\cite{Hwang2017CharacterlevelLM}, thus increasing the task complexity~\cite{clark-etal-2022-canine}.

\subsection{Evaluation}
Language models like mBERT~\cite{devlin-etal-2019-bert} and XLM-R~\cite{conneau-etal-2020-unsupervised} helped in a way to shift the overall focus towards a unified general multilingual framework of language modeling. With this advancement, researchers identified the need for a unified evaluation framework that can provide insight into the overall transfer-learning capability of a multilingual model. With this aim, in \cite{hu2020xtreme}, the authors designed an evaluation setting that ranges from syntactic to semantic tasks and structural prediction tasks. In each task evaluation, the training is performed in English, whereas the evaluation comprises 40 languages from 12 families, thus ensuring the typological diversity in language selection. Later, this practice of expanding over far-reach languages continued through subsequent studies~\cite{ruder-etal-2021-xtreme}. However, these unified evaluation frameworks still do not entirely comprise languages from highly low-resource languages and dialects.

\section{Open Problems}
\paragraph{Tokenization free approach}
In short, a silver-lined multilingual tokenizer is yet to be found as a monolingual subword tokenizer still outperforms a multilingual tokenizer in almost every task~\cite{rust-etal-2021-good,https://doi.org/10.48550/arxiv.2112.10508}. Now monolingual Training tokenizer every time is not a feasible option and the original multilingual tokenizer does not have vocabulary distributional knowledge about the unseen languages. One option could be the effort to rule out the tokenization step completely.
Several recent studies are exploring this direction. Models like CAINNE~\cite{clark-etal-2022-canine} are tokenization-free modules that encode texts differently. Instead of using a vocabulary and tokenization step, CANINE encodes texts at the character level, produces character-level output, and uses a soft-max layer-based smoothing for subword projection. Recently another model PIXEL~\cite{https://doi.org/10.48550/arxiv.2207.06991}, treats text as an image and by doing that, bypasses the text encoding step. These methods are getting inspiring results but the full potentiality of this direction is yet to unfold. For example, pixel receives impressive results for syntactic tasks, but this is not true for semantic tasks. Further studies need to be done on the applicability of tokenization-free models in the case of zero-shot and few-shot transfers. 

\paragraph{Extending to new languages}
In the current paradigm of MLLM, the curse-of-multilinguality~\cite{pfeiffer2022xmod} is an issue where the per-language performance drops unequally when the model is trained on multiple languages. In a general setting, the usual scenario is: (1) A monolingual model performs better than a multilingual model on specific tasks and language (2) The low-resourced ones perform poorly compared to the high-resourced ones in MLLM (3) Languages unseen during the pre-training stage perform the worst. To tackle this issue, the straightforward approach would be to finetune the model to the specific target language. However, then, the model becomes specific to only one language though it does not increase the catastrophic forgetting in general. The authors in \cite{muller-etal-2021-unseen} have identified then, while dealing with unseen language and scripts, having pretrained on any closely related language usually helps. Another approach to improve the capacity of MLLM would be to augment the vocabulary with new tokens, which works better for unseen languages and improve performance for the languages already seen during pre-training \cite{wang-etal-2020-extending}.
Another simple approach includes mitigating the cross-lingual transfer gap by just performing training on a bit of target data amount\cite{lauscher-etal-2020-zero}.
Another helpful approach to improve the model capacity is to use adapters, modular parameter units that can be injected in every layer of a base language model. These adapter layers can then be trained on language or task-specific data, thus enabling the cross-lingual transfer without changing the base model parameters\cite{pfeiffer-etal-2020-mad}. This approach can be further used for unseen language and optimized zero-shot transfer by using these adapters at the pre-training stage, which lifts the curse-of-multilingualism to a greater extent\cite{pfeiffer-etal-2022-lifting}.

\paragraph{Model Interpretability and Shared Representation}
It is hypothesized that most of the large neural models are over-parameterized~\cite{https://doi.org/10.48550/arxiv.1803.03635}, thus, resulting in unnecessary computation and storage costs. Effective model pruning could be a feasible approach to try in this regard. \cite{https://doi.org/10.48550/arxiv.2007.12223} is one of the earlier studies on pretrained bert with model pruning. The results demonstrate that the idea of lottery ticket observations~\cite{https://doi.org/10.48550/arxiv.1803.03635} (i.e., if we select multiple small networks, we will end up getting the similar performance of the larger model at one time) remain relevant in this context of language modeling also. We find matching subnetworks for a range of downstream tasks at 40\% to 90\% sparsity. Subsequently, \cite{prasanna-etal-2020-bert} has done LTH based empirical study on English bert. The findings suggest “random” subnetworks are still almost as good as the “good” ones, and even the “worst”(sampled elements which didn’t survive the pruning by LTH \cite{https://doi.org/10.48550/arxiv.1803.03635}) ones perform on par with a substantial baseline. \cite{https://doi.org/10.48550/arxiv.2205.12672} is one of the studies which explore LTH-based pruning in multilingual scenarios. The authors use the lottery ticket hypothesis in mBERT and conclude that the sub-networks found for different languages are similar. In addition, mBERT comprises a language-neutral sub-network shared among many languages, which is the most useful one while performing cross-lingual transfer. Another finding is for MLM tasks, the similar language \& task specific sub-network are primarily identical in the lower to the middle layer. In contrast, network similarity is visible mainly in the higher layer for NER \& XNLI tasks. The prospect of different pruning-based methods is further explored in \cite{https://doi.org/10.48550/arxiv.2204.02601}. The authors study two types of model pruning strategies: regularization-based and gradient-based pruning. They also propose a method called Dynamic Sparsification to allow training the model once and then adapting to different model sizes at inference. They also used a diversity loss to prune language-specific subnetworks. Results show that subnetworks of different languages are indeed different. The most straightforward pruning algorithm performs the best, and A fast model does not mean it should be small. Another finding is for large multilingual models like XLM-R sharing the subnetwork for a universal representation is preferable; the language-neutral part contributes the most in cross-lingual transfer, which also aligns with the finding of \cite{https://doi.org/10.48550/arxiv.2205.12672}. \cite{sajjad2021effect} did one study on the Effect of Dropping Layers of Pre-trained Transformer Models where they pruned BERT, Roberta, and XLNet models up to 40\% while maintaining up to 98\% of their actual performance. The findings are: (i) the lower layers are most critical to maintaining downstream task performance, (ii) some tasks, such as paraphrase detection and sentence similarity, are more robust to the dropping of layers, and (iii) models trained using a different objective function exhibit different learning patterns and w.r.t the layer dropping. 

Another question is whether MLLM learns universal patterns or not. Learning universal patterns is essential for effective cross-lingual transfer. From the discussion above, it can be inferred that MLLM contains particular language-specific and shared representation space. The shared representation space helps in probing-based tasks like POS tagging. However, performing complex tasks like MT is still beyond scope just by using this shared representation space.

Analysis of traditional classifier-based probing methods is another heavily investigated direction that is still being explored to ensure model interpretability. For example, \cite{https://doi.org/10.48550/arxiv.2205.02023} examines whether good networks include any superior linguistic knowledge or not. It results in not finding any interpretable patterns. \cite{10.1162/coli_a_00444} investigated Language Relationships in Multilingual Sentence Encoders Through the Lens of Linguistic Typology; This study looks into how languages are placed in multilingual subspace in mBERT and XLM-R. At the same time, how language specific sub-spaces within multilingual sentence encoders (LASER~\cite{Artetxe_2019}, m-BERT~\cite{devlin-etal-2019-bert}, XLM~\cite{https://doi.org/10.48550/arxiv.1901.07291}, and XLM-R~\cite{conneau-etal-2020-unsupervised}) concerning a range of typological properties pertaining to the lexical, morphological, and syntactic structure can be separated. Their results show interesting differences in encoding linguistic variation associated with different pre-training strategies. \cite{https://doi.org/10.48550/arxiv.2205.02023} is another recent study on linguistic probing using universal dependency tasks and data points.

\paragraph{Fairness in Language Models}
A substantial amount of work has investigated existing social bias (e.g., gender, racial, ethnic, occupational) identification and mitigation approach in PLMs, including reducing token sensitivity during text generation~\cite{https://doi.org/10.48550/arxiv.2106.13219}, investigating model sensitivity \cite{immer-etal-2022-probing}, prompting using natural sentences~\cite{alnegheimish-etal-2022-using} and probing via embedding lookup \cite{ahn-oh-2021-mitigating}. However, the state-of-the-art NLP models and datasets are still biased toward certain attributes~\cite{joshi-etal-2020-state}, and the overall utility provided by the models is still skewed~\cite{blasi-etal-2022-systematic}.





\section{Conclusion}
This survey provides insights into the current state of the art of multilingual text representation and the extent of language modeling. While many languages are already being covered under the current paradigm of language modeling, the full potential is not explored yet. There are languages left to cover. At the same time, the currently covered low-resourced ones are still not free from the impact of capacity dilution. Keeping this in focus, we did a review on the potential scope of improvement and research directions.

\ifCLASSOPTIONcaptionsoff
  \newpage
\fi

\bibliographystyle{IEEEtran}
\bibliography{IEEEexample}

\end{document}